\documentclass[10pt,twocolumn,letterpaper]{article}

\usepackage{times}
\usepackage{epsfig}
\usepackage{graphicx}
\usepackage{amsmath}
\usepackage{amssymb}
\usepackage{enumitem}
\usepackage{color}
\usepackage{rotating}
\usepackage{multirow}
\usepackage[english]{babel}
\usepackage{mathtools}
\usepackage{subfigure}

\date{}
\begin{document}

\title{Visual-Semantic Scene Understanding by Sharing Labels in a Context Network}

\author{Ishani Chakraborty\\
Rutgers University\\
New Jersey, USA 08854\\
{\tt\small ishanic@cs.rutgers.edu}
\and
Ahmed Elgammal\\
Rutgers University\\
New Jersey, USA 08854\\
{\tt\small elgammal@cs.rutgers.edu}
}

\maketitle
  
\begin{abstract}
We consider the problem of naming objects in complex, natural scenes containing widely varying object appearance and subtly different names. Informed by cognitive research, we propose an approach based on sharing context based object hypotheses between visual and lexical spaces. 
To this end, we present the Visual Semantic Integration Model (VSIM) that represents object labels as entities shared between semantic and visual contexts and infers a new image by updating labels through context switching. At the core of VSIM is a semantic Pachinko Allocation Model and a visual nearest neighbor Latent Dirichlet Allocation Model. For inference, we derive an iterative Data Augmentation algorithm that pools the label probabilities and maximizes the joint label posterior of an image. Our model surpasses the performance of state-of-art methods in several visual tasks on the challenging SUN09 dataset.


 

\end{abstract}
\section{Introduction}
The human visual system is expert at parsing a complex scene and naming objects within it. But how does a human mind navigate the complex visual layout of objects while using the lexical or semantic knowledge of the environment to precisely identify objects in the scene? While the exact mechanisms are yet unknown, \textit{sharing context based object hypotheses} across visual and lexical spaces is known to be one of the key guiding principles in cognition~\cite{moshebar, swinney}.

\begin{figure}[t]
\centering
\includegraphics[width=9cm]{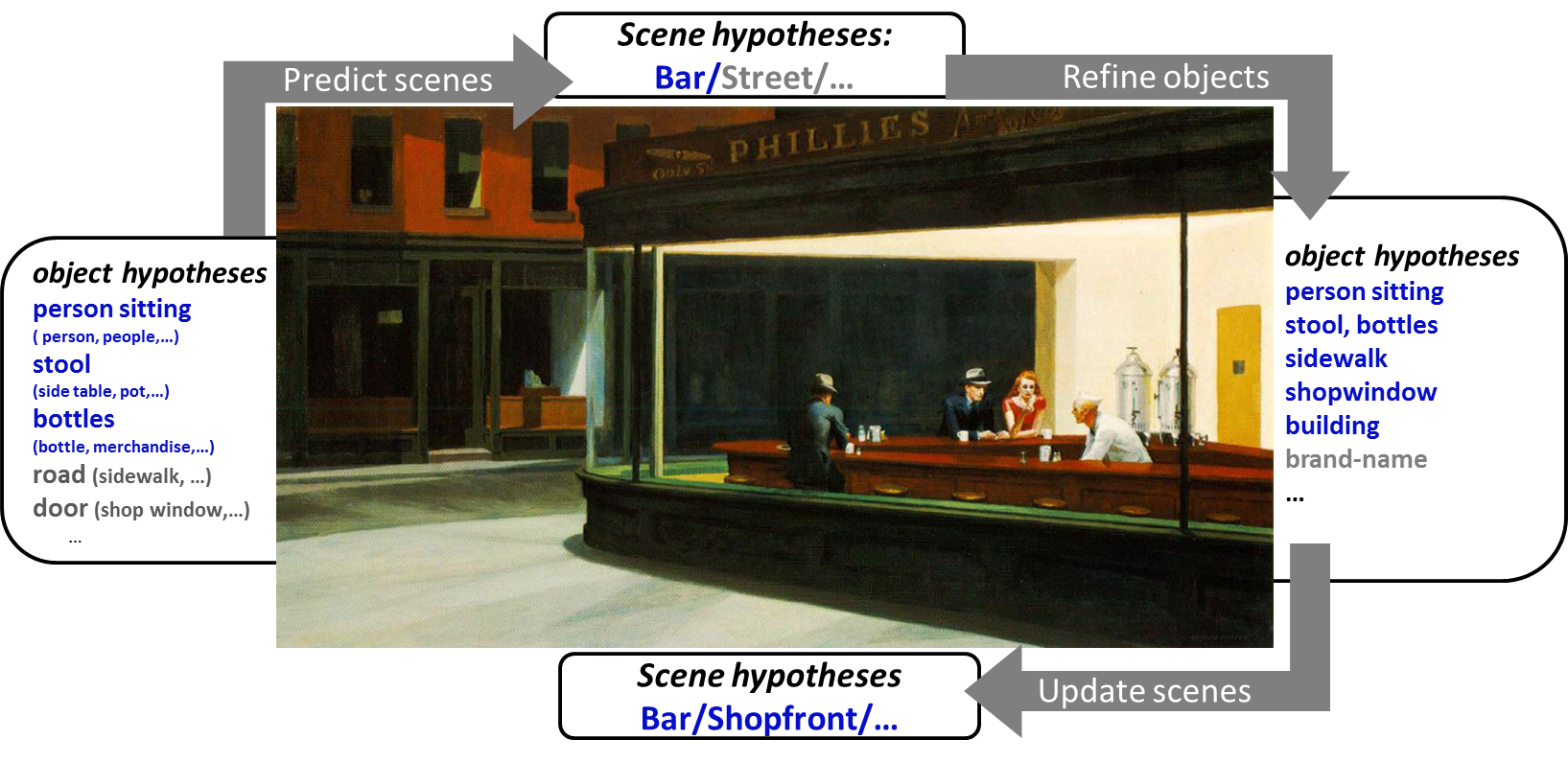}
\caption{Top-down facilitation of scene understanding by sharing labels through visual and semantic contexts.}
\label{fig:overview}
\end{figure}

Consider the famous Nighthawks painting in Figure~\ref{fig:overview}. At the first glance, the scene is hypothesized into individual objects such as a person, street, etc. Some of these objects cohere semantically to create a scene context of a roadside bar with people (and other possibly incorrect contexts). In effect, confidence in some objects can reduce, categories may become more specific (e.g. road as a sidewalk) and some new objects may appear (such as buildings, which may not be visually evident at the first glance). We repeat this process of updating our object hypotheses by iterating through our visual and semantic contextual knowledge base.

To facilitate this joint inference, we conceptualize scene understanding as a top-down integration of lexical and visual spaces via object names. The lexical space is the vocabulary of all object names in the knowledge base, while the visual space maps the visual appearances to object names. Each space has a different contextual relationship between objects. In the lexical space, related objects typically appear together in a given environment (semantic context), while in the visual space, related objects are visually similar in their appearance (visual context). 

To this end, we present the Visual Semantic Integration Model (VSIM) that connects the semantic and visual contexts through their shared object names. VSIM models scene interpretation as a top-down approach where semantically contextual labels are first created to represent a coherent scene composition. These labels are then re-interpreted with their visually contextual counterparts in the appearance space. Specifically, VSIM is a probabilistic, hierarchical model of latent context and observed features. In the first level, the image is modeled as a distribution over latent semantic contexts which determines the semantic labels that compose the scene. In the next level, each semantic label's visual context determines the appearance features that are finally the observed variables in the model. Inference in VSIM is initiated in a bottom-up manner, where observed image regions are the only cues used to infer the semantic and visual object labels in the image. I.e., the  goal of VSIM is to infer the semantic object labels in an image, given its appearance features. The general overview and our main contributions are as follows. 
\begin{itemize}
\item{For representing complex scene semantics, we introduce Pachinko Allocation Model (PAM) to effectively capture the semantic hierarchy of concepts in natural images through a directed graphical structure.} 
\item{For representing visual context, we propose nearest neighbor based Latent Dirichlet Allocation (nnLDA) that finds discriminative visual concepts. nnLDA exploits the strength of nearest neighbor decisions within a structured generative LDA approach.} 
\item{To infer labels in a new image, we derive an iterative Data Augmentation algorithm that alternates between the two context spaces to correctly pool the label probabilities inferred from each space and maximize the label posterior for the image.}
\item{Finally, our \textbf{Visual Semantic Integration Model (VSIM)} is motivated by the human cognitive process of shared context and represents a novel algorithmic formulation of that process. It mimics the cognitive process by representing object labels as entities shared between semantic and visual contexts and infering a new image by updating labels through context switching. This is the most significant contribution of this paper and conceptually different from previous approaches where context has been used mostly as a filter to reduce false detections.}
\end{itemize}
Our novel approach combined with an appropriate probabilistic technique for inference is able to surpass the state-of-the-art approaches for identifying diverse object categories in natural scenes.

\section{Related Work}
\textbf{Context in cognition} has been studied in psychophysics and linguistics. Particularly, studies by Bar et al.~\cite{moshebar} found evidence of an interactive context network in the brain that facilitates object prediction through the so-called \textit{context frames} of reference that bind visually or semantically related objects. Swinney's Cross-Modal Priming Task~\cite{swinney} proved that lexical access follows a \textit{multiple hypothesis} model where listeners accessed multiple meanings for ambiguous words even when faced with strong biasing contexts. These findings provide a strong motivation towards modeling an interactive context network integrating visual and lexical spaces. 

\textbf{Mapping images to related text} is gaining importance in large scale learning of web images. One strand of research is aimed at generating natural language sentences from objects and their inter-relations~\cite{tamara}. Our problem is related to joint image and word sense discrimination encountered in image retrieval tasks. These works have analyzed polysemy in images returned from keyword searches, in terms of visual senses of keywords. However the ambiguity in these tasks lies mostly in the visual domain since keywords are usually static, sparse and well-defined. Hence, the \textit{sense mapping} between keywords and images is either abstracted through a single latent sense~\cite{wsd}, picked up from knowledge sources e.g., wikipedia or the image and text words are jointly modeled through a single latent variable~\cite{corrlda}. As shown in the results, these simple correlations are not effective in mapping the rich interactions between semantic and visual space.  

\textbf{Hierarchical context networks} provide a nice framework for scene understanding due to the modular separation of concepts at different granularities. Mostly, previous work has used semantic networks as filters to remove incompatible object detections in the scene~\cite{hcontext, crf}. A visual hierarchy of object classes is proposed in~\cite{discriminativehierarchy_koller}. Our work is related to the topic modeling algorithms for scene understanding~\cite{sudderth,sltm,corrlda,tsu}. However, these models try to capture overlapping information between images and text to reinforce each other. In contrast, our method captures the complementary information in these contexts and exploits them to improve the quality of the inferred labels. To the best of our knowledge, no previous work has considered such joint inference framework across dichotomous information spaces.

\begin{figure}[t]
\centering
\includegraphics[width=8cm]{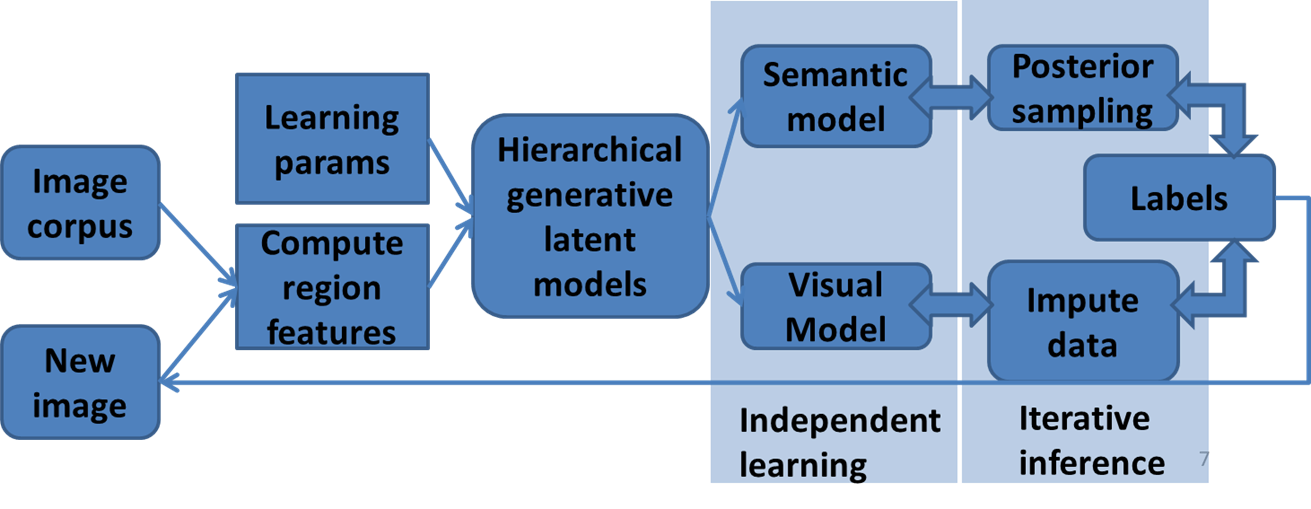}
\caption{Flowchart of the overall approach.}
\label{fig:flowchart}
\end{figure}

\begin{figure*}[t]
\centering
\begin{minipage}[b]{0.45\linewidth}
\centering
\includegraphics[width=\textwidth, height = 4.5cm]{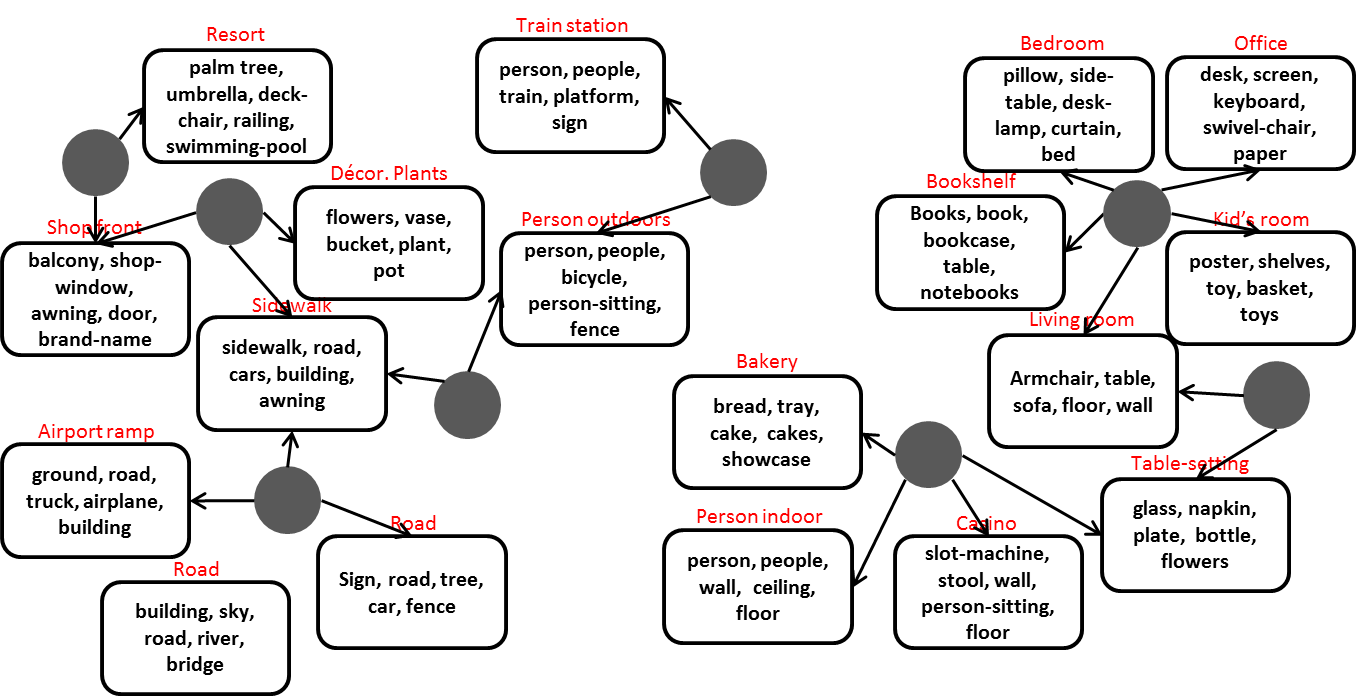}
\caption{Part of the semantic context graph. Supertopics (gray nodes) and subtopics (shown by most frequent labels). 
Our interpretation of each subtopic is denoted in red.}
\label{fig:pam}
\end{minipage}
\hspace{0.5cm}
\begin{minipage}[b]{0.45\linewidth}
\centering
\includegraphics[width=\textwidth, height = 4.5cm]{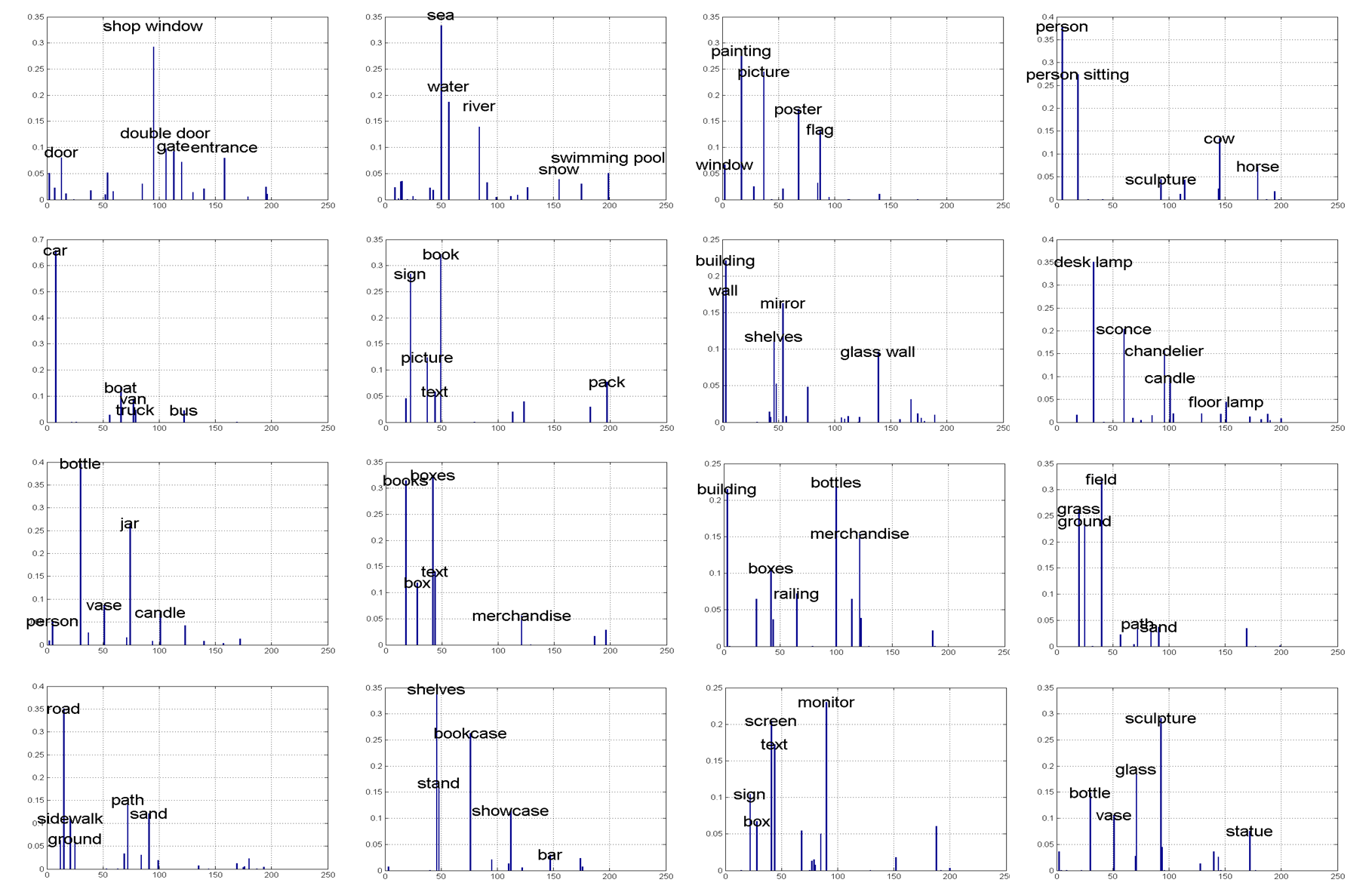}
\caption{Some of the visual topics. The top 5 labels are marked. Some clusters capture general to specific objects (door, shop window), casual coincidences (book, text) and intra-class variabilities (person,animals and person,bottles)}
\label{fig:nnlda}
\end{minipage}
\end{figure*}

\section{Modeling Context Network}
\label{sec:model_context}
Given an image, we wish to predict a set of objects that best fit the image content. VSIM models these object labels as connection between two different context networks. The semantic context of labels is modeled in a Pachinko Allocation model (PAM) through a hierarchy of semantic supertopics and subtopics. The visual context of labels is established through visual topics of a nearest neighbor Latent Dirichlet Allocation (nnLDA). Intuitively, the topic distributions encode the grouping between labels in these two contexts. We briefly give an overview of our approach. 
\begin{figure*}[t]
\centering
\includegraphics[width=16cm]{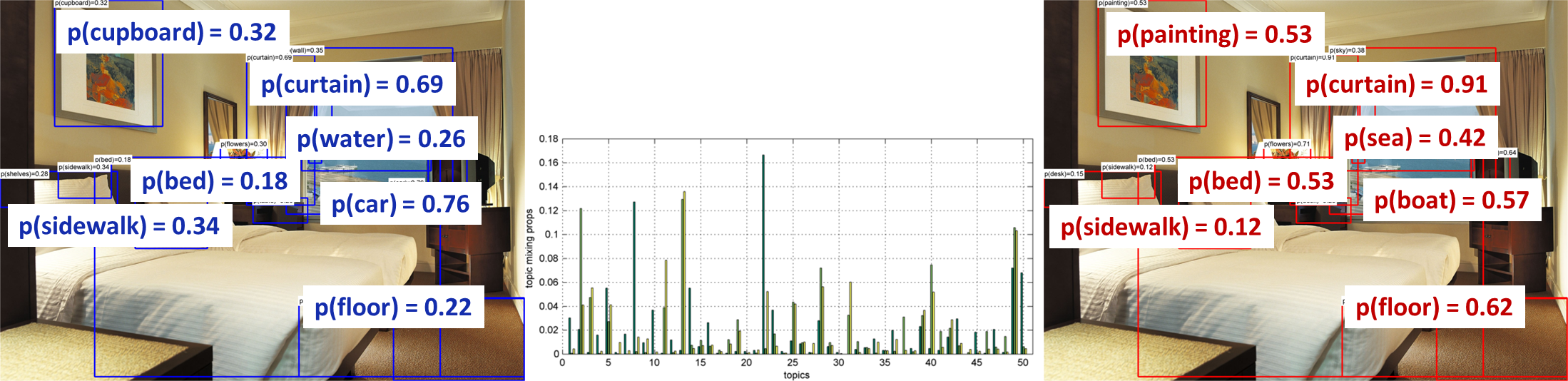}
\caption{Left: predicted object labels using visual context alone (initial). Right: Maximum aposteriori labels after joint inference (final). Middle: initial (green) and final (yellow) predicted semantic subtopic distribution.}
\label{fig:roomwithaview}
\end{figure*}
\subsection{Semantic context in lexical space} 
A complex natural scene may be composed of multiple sub-scenes each with distinct coherency of objects within them. Thus, scene context should be established through a hierarchy of semantics. Single level topic models like Latent Dirichlet Allocation which is commonly used for modeling context in images~\cite{sudderth,corrlda} cannot encode such relationships. In VSIM, this complexity of semantic context is encoded with a probabilistic Directed Acyclic Graph (DAG) of topics known as Pachinko Allocation Model (PAM). Unlike the single level LDA models, PAM explicitly models relations among words and topics through arbitrary, nested and possibly sparse dependencies. In natural scenes, this enables discovery of fine-grained and tightly coherent subscenes. 

Figure~\ref{fig:pam} shows a subset of the semantic context network of supertopics and subtopics. These topics are learnt from co-occurrence statistics of object labels in images from SUN09 dataset. Not only do labels that occur together very frequently form a strong clusters in subtopic space (bookshelf subtopic: books, notebook, table), but also the related subtopics are learnt as a higher level supertopic (a bookshelf can be found in isolation or can occur with a living room). Without such a explicit hierarchy of topic ontologies encoded in the PAM, such relations would get captured as ``nonsensical" topics. 
\begin{figure}[h]
\centering
\includegraphics[width=5cm]{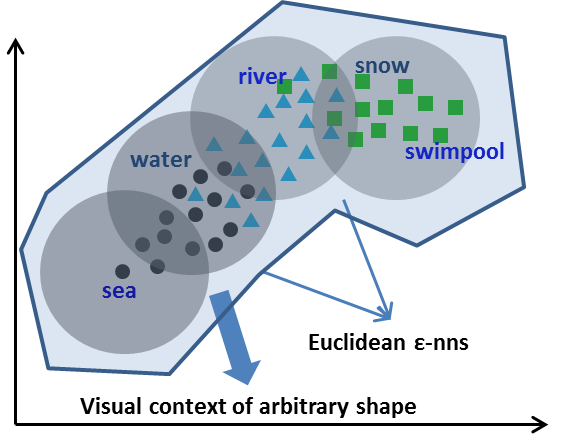}
\caption{Illustration of discovery of visual topic manifold of \{sea, river, snow, water, swimpool\} in nnLDA. While nearest neighbors capture dense matchings, a topic manifold captures implicit, spatially extended and sparse relations between labels.} 
\label{fig:knntopic}
\end{figure}

\subsection{Appearance context in visual space} 
Most latent variable models quantize rich image features to visual words to facilitate a multinomial (word count) modeling of image data. To avoid this lossy quantization while keeping the convenience of a multinomial inference, we represent images in a supervised feature space using a \textit{bag-of-labels} formulation. Here, the feature space is spatially clustered using nearest neighbor groupings of features and a bag-of-labels is constructed from each grouping. 
Such a nearest neighborhood is effective in finding dense similarities within a geometrically constrained location. However, it does not capture the implicit, spatially extended and sparse relations between labels. To learn such relations, we construct bag-of-labels for each region and perform LDA on it. This topic model formulation enables the rich feature space to be projected into arbitrary topic manifolds (Figure~\ref{fig:knntopic}), such that sparse and strong visual similarity in labels can be discovered. Some other topic distributions are shown in Figure~\ref{fig:nnlda} 

\subsection{Inference by label sharing between switching contexts} 
Labels are inferred in VSIM through joint inference in the semantic and visual space. We derive an iterative Data Augmentation algorithm which alternates between the two spaces to arrive at the joint inference. The \textit{room with a view} (Figure~\ref{fig:roomwithaview}) illustrates the functioning of our algorithm. The left panel shows some regions along with an initial set of most-likely labels using visual context alone. The right panel shows the final, maximum aposteriori (MAP) labels of the same regions after joint inference. The middle panel shows the initial (green) and final (yellow) distributions over inferred semantic subtopics. The joint inference shows a behavior quite similar to the cognitive process described in the introduction. Some MAP labels show an increase in beliefs, such as "bed", "cushion", and "curtain". Some labels are updated to a more \textit{specific} class e.g., "water" is relabeled as "sea". Ambiguity between visually similar labels is reduced as contextually more appropriate labels are enhanced. For example, ``car" is changed to ``boat", since ``boat" is visually similar to ``car" but fits better with ``sea". The label probabilities at regions which don't fit any context become more diffused, e.g., p(sidewalk) and hence can be thresholded out. As label probabilities at image regions converge, the semantic subtopic distribution becomes more peaky. By alternating between the two spaces, the scene finally converges to bedroom and seaview related concepts (topic 11 and 13, resp.).
\begin{figure*}[t]
\centering
\begin{minipage}[b]{0.45\linewidth}
\centering
\includegraphics[width=7.8cm]{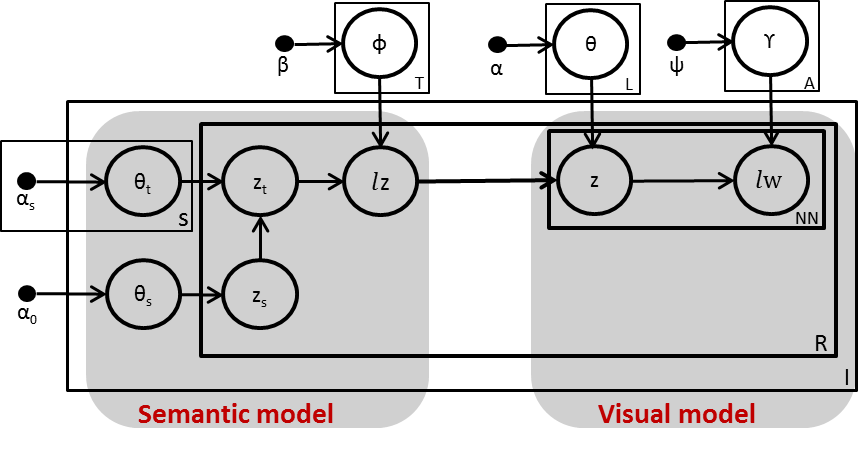}
\caption{Visual-Semantic Integration Model (VSIM)}
\label{fig:vsim}
\end{minipage}
\hspace{0.5cm}
\small
\begin{minipage}[b]{0.45\linewidth}
\centering
\begin{tabular}{|l|l|}\hline
$I$ 		& number of images in the corpus \\
\hline
$R$ 		& variable number of regions per image \\
\hline
$NN$ 		& number of $\epsilon$-nearest labels of an image region \\
\hline
$S$		& number of semantic supertopics \\
\hline
$T$		& number of semantic subtopics \\
\hline
$A$		& number of visual topics \\
\hline
$L$		& number of object labels \\
\hline
$z_s$,$z_t$,z		& semantic super, subtopic and visual topic sample \\
\hline
$lz$,$lw$		& semantic and observed label sample \\
\hline
($\alpha_0$,$\theta_s$)		& $\mathit{Dir-Mult}$ of supertopics, per image\\
\hline
($\alpha_s$,$\theta_t$)		& $\mathit{Dir-Mult}$ of subtopics, per image\\
\hline
($\beta$,$\phi$)		& $\mathit{Dir-Mult}$ over semantic labels, per subtopic \\
\hline
($\alpha$,$\theta$)		& $\mathit{Dir-Mult}$ of visual topics, per object label \\
\hline
($\psi$,$\gamma$)		& $\mathit{Dir-Mult}$ over labels, per visual topic \\
\hline
\end{tabular}
\caption{Definition of variables in VSIM}  
\label{tab:results}
\end{minipage}
\end{figure*}
\normalsize

\section{The Visual-Semantic Integration Model}
\label{sec:vsim}

Fig.~\ref{fig:vsim} shows the plate notation of the VSIM graphical model. The core of VSIM is a cascade of two models: the PAM model which generates semantic labels, followed by the nnLDA model which generates the observed labels. Context is represented through topics, which are Dirichlet-Multinomial distributions over labels. Hence, a semantic topic is learnt as a probabilistic group of labels that co-occur frequently in images. A visual topic is learnt when labels map to similar appearance features and thus get grouped frequently in different bags of labels. Finally, the structural hierarchy of supertopics and subtopics in the PAM are also probabilistically estimated by adapting the scale parameters of the subtopic Dirichlets based on the most likely paths discovered by the model.      

For inference, we derive an iterative Data Augmentation (DA) algorithm that alternates between the two context spaces to correctly pool the label probabilities inferred in each space and maximizes the label posteriors. Concisely, the inference is seeded with the most likely labels based on image features alone. This is achieved by passing the bag of labels through nnLDA inference. Within each iteration, label samples are drawn from the current distribution and used to estimate scene multinomials in the PAM. Then, these semantic multinomials update the label distribution. This information is propagated back to the visual model to update and normalize the observed labels at each image location. We use \textit{collapsed Gibbs sampling} for estimating topic distributions in each iteration. In the following, we provide details of parameter estimation and inference in VSIM.

\subsection{Semantic Context: Generating object labels.}
\label{sec:scene_context_model}

We model the co-occurrence context of object labels using a three-level Pachinko Allocation Model (PAM)~\cite{pam}. Given an image corpus of size $I$, labels $l$ of an image $d$ are generated by sampling topics at two levels. The per-image supertopic multinomials, ${\theta}s$ are sampled from a symmetric Dirichlet hyperparameter $\alpha_0$, while subtopic multinomials, ${\theta}t$ are sampled for each supertopic from an asymmetric Dirichlet hyperparameter $\alpha_s$. The role of $\alpha_s$ is crucial since it establishes sparse DAG structure between super and subtopics. The label mixing multinomials $\phi$ per subtopic are sampled corpus-wide from a symmetric Dirichlet hyperparameter $\beta$. Finally, each label $l$ in the image is sampled from a topic path $(zs,zt)$. We refer to these object labels as the \textit{semantic labels}. 
\begin{itemize} 
\item{For each image $d = \{1\cdots I\}$, ${\theta}s \sim \mathit{Dir}(\alpha_0), {\theta}t \sim \mathit{Dir}(\alpha_s), s = \{1\cdots S\}$.}
\item{For each subtopic $zt = \{1\cdots T\}$, $\phi_t \sim \mathit{Dir}(\beta)$.}
\item{For each label $lz = \{1\cdots L\}$ - }
\begin{itemize}
\item{Sample a topic path: $zs \sim \mathit{Mult}({\theta}s), zt \sim \mathit{Mult}({\theta}t_{zs}$).}
\item{Sample a label from subtopic, $lz = \mathit{Mult}(\phi_{zt})$.}
\end{itemize}
\end{itemize}

\subsection{Appearance Context: Generating image regions}
\label{sec:app_context_model}

We formulate appearance context through a bag-of-labels representation. To achieve this, we first project image regions and their corresponding labels into a supervised feature space and find their nearest labels in an $\epsilon$ neighborhood. Thus, each groundtruth (semantic label) corresponds to a bag of observed labels based on its image features. Let $\{\vec{f_1}, \vec{f_2},\cdots\,\vec{f_r}\}$ be the features of $r$ image regions in a feature space  (e.g., SIFT) with the corresponding semantic labels $\{lz_1, lz_2,\cdots\,lz_r\}$. The bag-of-labels is computed as follows. 
\begin{equation}
lw_r = \{l\sp{\prime} \mid \; \forall j; \parallel f_r(lz_r)-f_j({l\sp{\prime}})\parallel \leq \epsilon\},
\end{equation}
where, $\parallel.\parallel$ is a distance norm in the feature space. Thus, a semantic label $lz$ is associated with a set of observed labels $lw = \{l\sp{\prime}\}$. 
This induces a many-to-many, bipartite relation between semantic labels and the observed labels (from the same label pool), which is then modeled effectively using an LDA. Specifically, the topic multinomials, $\theta$ capture the visual polysemous relation between one semantic label $lz$ and multiple topics while the label mixing multinomials $\gamma$ capture the visual synonymous relation between multiple topics and one observed label $lw$. The generative process is as follows.
\begin{itemize}
\item{For each label $lz = \{1\cdots L\}$, $\theta_l \sim \mathit{Dir}(\alpha)$.}
\item{For each visual topic $z = \{1\cdots A\}$, $\gamma_z \sim \mathit{Dir}(\psi)$}
\item{For each of the $NN$ observed labels}
\begin{itemize}
\item{Sample a visual topic $z$, $z \sim \mathit{Mult}(\theta_{lz})$}
\item{Sample a label $lw = \mathit{Mult}(\gamma_z)$}.
\end{itemize}
\end{itemize}

\subsection{Parameter Learning and Inference}
The joint probability distribution over all the variables in the model, given hyperparameters:
\begin{align}
J.P.D. & = Pr( \overbrace{\textstyle zs,zt,lz,z}^{\mathclap{\text{hidden variables}}},\underbrace{\textstyle lw}_{\mathclap{\text{observables}}},\overbrace{\textstyle \alpha_s,{\theta}s,{\theta}t,\phi,\theta,\gamma}^{\mathclap{\text{parameters}}} )
\end{align}
For learning the parameters in the model, we use annotated images, hence $lz$ is known. We use these groundtruth labels to ground the semantic labels during learning due to which the semantic and the visual models become conditionally independent. We use collapsed Gibbs sampling for estimating topic distributions in each context space.
\subsection{Estimating semantic topics}
The joint distribution of the semantic model from the above J.P.D. formulation yields:
\begin{align}
JPD_{semantic} & = \{\displaystyle\prod_I p({\theta}s|\alpha_0) (\displaystyle\prod_S p({\theta}t|\alpha_s)) \\ \nonumber
& \displaystyle\prod_R p(zs|{\theta}s) p(zt|{\theta}s_{zs}) p(lz|\phi_{zt}) \displaystyle\prod_T p(\phi|\beta)\} 
\end{align}
The proposal for Gibbs sampling of supertopic and subtopic pairs for the $i^{th}$ label is derived to be~\cite{pam}:
\small
\begin{align}
& P(zs_i = s, zt_i = t | lz = l, zs_{\sim i}, zt_{\sim i},\alpha_0,{\alpha}s_t,\beta)  \propto \\ 
& \left(\frac{n^d_s + \alpha_0}{\displaystyle\sum\limits_S n_s^d + S\alpha_0}\right) 
	\left(\frac{n^d_{st} + {\alpha}s_{st}}{\displaystyle\sum\limits_T n_{st}^{d} + \displaystyle\sum\limits_T {\alpha}s_{st}}\right)
	\left(\frac{n^t_l + \beta}{\displaystyle\sum\limits_L n_l^t + L\beta}\right), 
\end{align}
\normalsize
where $zs_i$ and $zt_i$ are supertopic and subtopic assignments for $lz_i$, and $zs_{\sim i}$ and $zt_{\sim i}$ are topic assignments for all the remaining labels in the image. Excluding the current token, $n^d_s$ is the count of topic $s$ in image $d$ and $n^d_{st}$ is the number of times subtopic $t$ is sampled from supertopic $s$ within image $d$. $n^t_l$ denotes the number of times label $l$ is assigned to subtopic $t$ in the entire corpus.

\subsection{Estimating topic hierarchy} 
We estimate ${\alpha}s$ within each Gibbs iteration of the PAM. These hyperparameter values capture the structural links between supertopics and subtopics. Therefore, the strength of the these connections need to be estimated in a data-driven manner. We use co-occurrence counts of super and subtopics to estimate ${\alpha}s$. Specifically, we use moment matching to estimate the approximate MLE of ${\alpha}s$. In this technique, the model mean and variance of each ${\alpha}s_{st}$ is computed by matching them to the sample mean and variance of topics' co-occurrence counts across all images.
\footnotesize
\begin{align} \label{eq:alpha_est}
E[{\alpha}s_{st}] & = \frac{{\alpha}s_{st}}{\displaystyle\sum\limits_T {\alpha}s_{st}} \nonumber
= \frac{{\alpha}s_{st}}{exp(log\sum_T{\alpha}s_{st})} \nonumber
= \frac{1}{N}\displaystyle\sum\limits_{I}\frac{n^d_{st}}{\displaystyle\sum\limits_{T} n^d_{st}} \\ \nonumber
log\sum_T{\alpha}s_k & = \frac{1}{T-1}\displaystyle\sum\limits_{T-1}log\left(\frac{E[{\alpha}s_{st}](1-E[{\alpha}s_{st}])}{var[{\alpha}s_{st}]}-1\right) \\ 
\end{align}
\normalsize

\subsection{Estimating visual topics} 
Starting with the joint distribution of visual model, the proposal distribution for visual topic of the $i^{th}$ label is derived to be:
\footnotesize
\begin{align}
& JPD_{visual} = \displaystyle\prod_{NN} p(z|\theta_{lz}) p(lw|\gamma_z) \nonumber
\{\displaystyle\prod_L p(\theta_l|\alpha)\} \{\displaystyle\prod_A p(\gamma_a|\psi)\} \\ \nonumber
& P(z_i = a | lw, lz, z_{\sim i}, \alpha, \psi)  \propto \nonumber
\left(\frac{n^{lz}_a + \alpha}{\displaystyle\sum\limits_A n^{lz}_a + A\alpha}\right) \nonumber 
	\left(\frac{n^a_{lw} + \beta}{\displaystyle\sum\limits_L n^a_{lw} + L\beta}\right), \\
\end{align}
\normalsize
where $z_i$ is the visual topic of the $i^{th}$ semantic label, $n^{lz}_a$ is the number of times topic $a$ is sampled for semantic label $lz$. $n^a_{lw}$ denotes the count when an observed label $lw$ is assigned to topic $a$ across the entire corpus. To get an intuitive insight into the counts that relate topics and labels, we consider a pair of labels $(lz, lw)$ and a topic $a$. If both $n^{lz}_a$ and $n^a_{lw}$ are low, topics would be assigned at random. If $n^{lz}_a$ is high but $n^a_{lw}$ is low it means that topic $a$ is consistent with $lz$ but the observed $lw$ is an outlier. If $n^{lz}_a$ is low but $n^a_{lw}$ is high, the observed label has a generic appearance (e.g., a white wall) so it is matched to many objects. The signal is relevant and peaky only when $n^{lz}_a$ and $n^a_{lw}$ are both high, which implies that topic $a$ would be consistently sampled for this $(lz, lw)$ pair and they would be grouped together.


\subsubsection{Inference}

Given an image, VSIM model needs to compute posterior probabilities over the semantic labels $lz$ for each image region, conditioned on the bag of observed labels $lw$. This distribution containing latent variables is as follows. 
\begin{align}
P(lz|lw) & = \sum_{zs,zt,z} P(lz,zs,zt,z|lw) \\ \nonumber
& = \sum_{zs,zt,z} P(lz|zs,zt,z,lw) P(zs,zt,z|lw) 
\end{align}
In the second equation, the first term denotes the conditional probability of $lz$ given augmented (observed and latent variables) data $(zs,zt,z,lw)$. The second term gives the predictive likelihood of latent data given observations. Based on the dependencies from the graphical model, it can be computed by marginalizing over $lz$, as follows. 
\begin{align}
P(zs,zt,z|lw) = \sum_{lz} P(zs,zt|lz) P(lz|lw) P(z|lw,lz) 
\end{align}
The above formulation leads to a coupled inference problem. For solving $P(lz|lw)$, we need the predictive topic probabilities $P(zs,zt,a|lw)$. However, since the link between the semantic topics and the observed $lw$ passes through $lz$, it needs to be marginalized out. 

We derive a Data Augmentation algorithm~\cite{tanner} to solve this inference problem. The idea of DA is similar to Expectation Maximization, but applies to posterior sampling. The general framework consists of an iterative sampling framework with two steps (1) \textit{Data imputation step}, in which the current guess of the posterior distribution $p(lz|lw)$ is used to generate multiple samples of the hidden variables $(zs, zt, z)$ from the predictive distribution in Eq. 9 and (2) \textit{Posterior sampling step}, in which the posterior is updated to be a mixture of the $N^s$ augmented posteriors and approximated to be the average of $p(lz| lw, zs, zt, a)$. Thus, a stationary posterior distribution is achieved through successive substitution. Formally, the two steps can be represented as:
\subsection{Data Imputation}
\footnotesize
\begin{align}
\{zs^{(t+1)}, zt^{(t+1)}, z^{(t+1)}\} \sim \sum_{lz} P(zs,zt|lz) P(z|lw, lz) P^{(t)}(lz|lw)
\end{align}
\normalsize
In this step, we begin from the visual end of the model. We create bag of labels for each image region and perform nnLDA inference. The Gibbs proposal during inference updates only the topic assignments of the new labels, while keeping the counts obtained from the learning phase fixed. The $\theta^r$ for a region is computed after a number of iterations of topic sampling is completed (100, in our case).   
\footnotesize
\begin{align}
p(\tilde{a}|lw,\tilde{a}_{\sim i}; \mathcal{M}) =  & (\frac{\tilde{n}^{r}_{a,\sim i} + \alpha}{\displaystyle\sum\limits_A \tilde{n}^{r}_{a} + A\alpha}) 
	(\frac{n^a_{lw} + \tilde{n}^a_{lw,\sim i} + \beta}{\displaystyle\sum\limits_L n^a_{lw} + + \tilde{n}^a_{lw,\sim i} + L\beta}), \\
P(\tilde{a}|lw) & = \tilde{\theta}^r_a = \frac{\tilde{n}^r_a + \alpha_a}{\sum_A \tilde{n}^r_a + \alpha_a}
\end{align}
\normalsize
Using the estimated topic proportions, we compute the likelihood of $lz$ at each region:
\footnotesize
\begin{align}
P(lz|lw) & = \sum_A P(lz=l|z=a)P(z=a|lw) \\ \nonumber
& = \sum_A \frac{P(z=a|lz=l)P(lz=l)}{P(z=a)}P(z=a|lw) = \theta^l_a \frac{n_l}{n_a}\tilde{\theta}^r_a,
\end{align}
\normalsize
$\theta_a^l$ is the learnt topic proportions for label $l$ and $\tilde{\theta}^r_a$ is the estimated topic proportions for the new image region $r$. $n_l$ and $n_a$ are the corpus wide counts of label $l$ and topic $a$, resp. From this multinomial distribution, we now draw samples of $lz$. These are used as observations for the semantic model. 
The Gibbs proposal for inference in PAM is similar in form to Eq. 4 (estimation proposal for sampling $(zs,zt)$), however only the topic assignments for new labels are changed (as in nnLDA inference). (See supplemental for the derivation). Thus, for each image, we are able to compute a complete set of topic samples $(zs, zt, z)$.  

\subsection{Posterior Sampling}
\small
\begin{align}
P^{(t+1)}(lz|lw) \sim \frac{1}{N^s} \sum_{N^s} P(lz|zs^{(t+1)},zt^{(t+1)},\mathcal{M})\cdot P^{(t)}(lw)
\end{align}
\normalsize
In this step, we start from the semantic end of the model. The imputed topic samples alongwith the learnt semantic parameters $\{{\tilde{\theta}s},{\tilde{\theta}t},\phi\}$ are used to generatively draw the semantic labels for an image. The average over multiple samples $N^s$ is used to update the $P(lz|lw)$ distribution. This new semantic label distribution is used to modulate the observed label distribution at each image region. After this step, we return to data imputation to use the new set of observed label probabilities.

\section{Experiments}
\label{sec:experiments}
\textbf{Dataset and Experiment Settings:} We evaluate our proposed VSIM model by performing  different visual tasks on the SUN09 dataset~\cite{hcontext}. SUN09 is a collection of 8600 natural, indoor and outdoor images. Each image contains an average of 7 different annotated objects and the average occupancy of each object is 5\% of image size. The frequencies of object categories follow a power law distribution. We consider the top 200 categories. 
4367 images were considered for learning the models and 4317 images were used for testing. For learning the model, we use the annotated ground-truth locations and labels provided with the dataset. In test images, we use the bounding boxes detected by DPM~\cite{dpm} detector as image regions, but not their decisions. A $256 \times 256$ image has about 400 regions.

\textbf{Feature Representation:} Each image region is represented using three types of features, as described in~\cite{tomasz}. Color is represented by normalized R, G, B histograms with their means and variances for a 36 length vector. Texture is captured using a 40-filter bank textons. We use a codebook of 100 textons for a texton histogram. Dense SIFT features are used for discriminative patterns using 400 word histogram. Each feature space is used to generate a bag-of-labels representation that feeds into the visual context model.

\textbf{Model representation:} The PAM is learnt with 20 supertopics and 50 subtopics. The supertopic Dirichlet $\alpha_0$ is set to a uniform value of 1. The subtopic Dirichlet ${\alpha}s$ is learnt during parameter estimation. For nnLDA, we choose a neighborhood radius empirically for each feature space and 50 visual topics are learnt. During parameter estimation, the Gibbs sampling is run for 1000 iterations in each model. For posterior inference of topics, we use 100 iterations. During imputation step, 500 samples are generated for the average distribution. The DA algorithm is run for 6 iterations. The final label posterior distributions are thresholded for label retrieval. 

\subsection{Initial label prediction: nnLDA versus nearest neighbors (NN):}
To compare nnLDA with NN, we use average precision gain (AP gain) in label retrieval, in Table \ref{tab:results}. The mean AP gain across 200 object categories is $4.0\%$, in which 168 objects show positive gain. It is interesting to note that the objects with maximum gain are categories with few training examples in the dataset  e.g., pillow, text and are visually similar to frequent categories. In contrast, the objects with maximum loss are categories with distinct appearances e.g., shoes, ingots, which might be losing distinctiveness through contextual groupings. The results highlight that nnLDA is better at handling the data imbalance problem and for visually ambiguous objects.  
\begin{table}[t]
\centering
\begin{tabular}{|r|c||c|}\hline
 \multirow{8}{*}{\begin{sideways}Object classes\end{sideways}} &\textit{Highest AP gain} &\textit{Least AP gain} \\ 
 \cline{2-3}
 \cline{2-3}
   &pillow ({$\mathit+29.47$}) 		&shoes ({$\mathit-40.95$}) \\
\cline{2-3}
   &text ({$\mathit+15.37$})		&ingots ({$\mathit-13.19$})\\
\cline{2-3}
   &desk ({$\mathit+14.83$})		&fish ({$\mathit-8.80$})\\
\cline{2-3}
   &armchair ({$\mathit+12.55$})	&chandelier ({$\mathit-7.30$})\\
\cline{2-3}
   &flowers ({$\mathit+12.45$})	&monitor ({$\mathit-6.30$})\\
\cline{2-3}
   &cabinet ({$\mathit+12.01$})	&glass ({$\mathit-6.27$})\\
\cline{2-3}
   &fence ({$\mathit+11.19$})		&faucet ({$\mathit-6.06$})\\   
\hline
\multicolumn{3}{|c|}{Mean Average Precision improvement = $+4.19\%$}\\
\hline
\hline
\end{tabular}
\caption{Average precision improvement with nnLDA compared to NN.}  
\label{tab:results}
\end{table}

\subsection{Semantic scene prediction using VSIM:}
We compare the scene detection performance of our model vis-a-vis the groundtruth. The ground-truth scene multinomial is computed by grounding semantic labels with groundtruth labels and inferring the PAM super and subtopics. We estimate the multinomials from image regions. Symmetric Kullback-Leibler Divergence between subtopic multinomials is use to evaluate how closely our joint inference fits the true distribution.  

We also compare our performance to two baselines: Correspondence LDA (CorrLDA~\cite{corrlda}) and Total Scene Understanding model (TSU~\cite{tsu}). CorrLDA models both visual words and lexical words as children of the same topic. This implies that lexical labels and visual features must display similar contextual groupings. TSU models a single semantic topic for an image and assumes one-to-one correspondence between object labels and visual words. We implemented the supervised versions of both these models and compare with our performance.
The KL divergence measures is lowest for the joint VSIM model (Table 2). Conceptually, this means our model accurately maps the visual-semantic space and therefore generalizes better in the test set. Algorithmically, it implies that labels predicted by our joint inference technique closely match the groundtruth labels.  

\subsection{Predicting top labels:} 
\begin{figure*}[t]
\begin{minipage}[b]{0.45\linewidth}
\centering
\includegraphics[width=\textwidth]{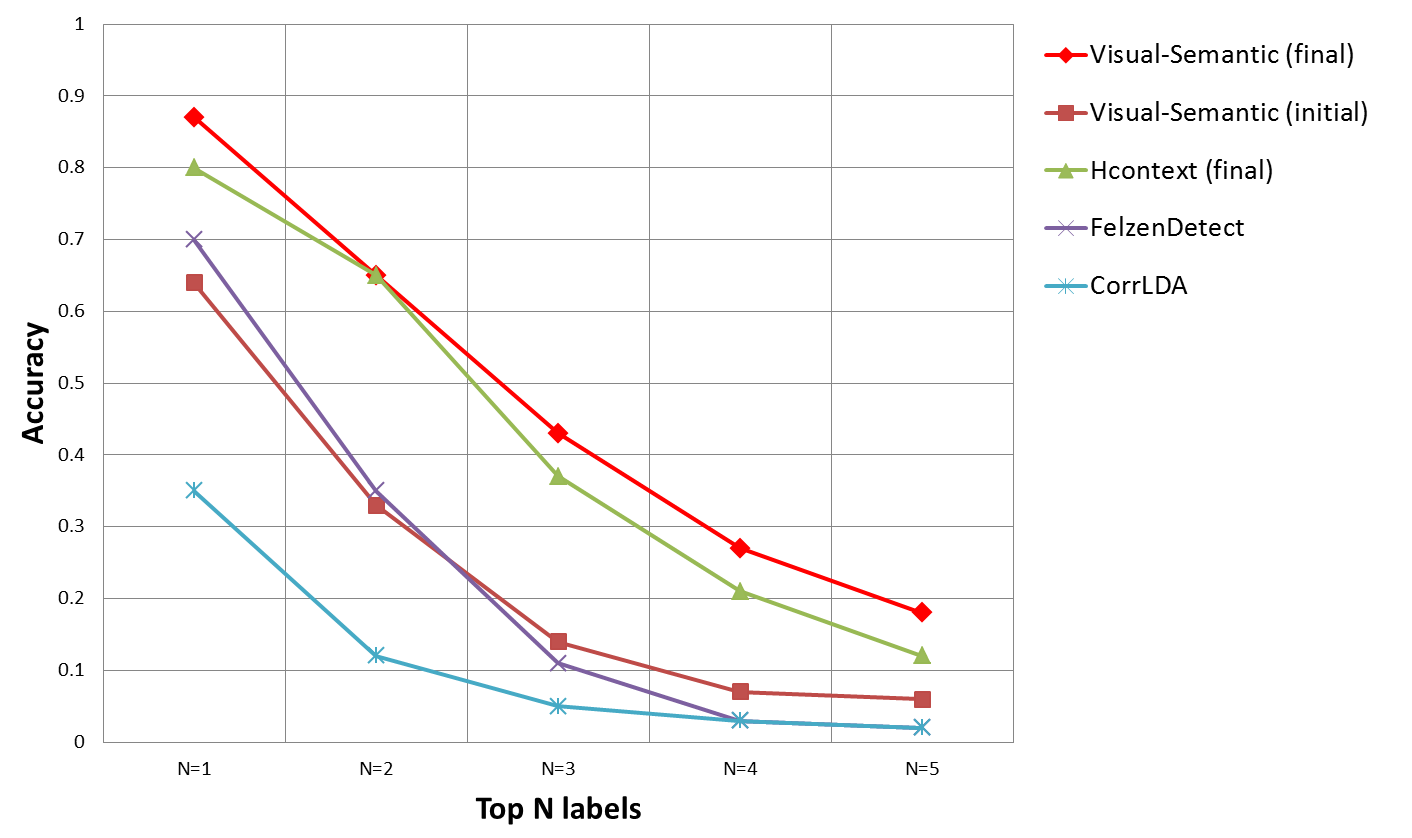}
\caption{Accuracy of prediction of top N (1-5) labels using VSIM on 200 categories versus hcontext on 107 categories and CorrLDA~\cite{corrlda}}
\label{fig:toplabels}
\end{minipage}
\hspace{0.5cm}
\begin{minipage}[b]{0.45\linewidth}
\centering
\includegraphics[width=\textwidth]{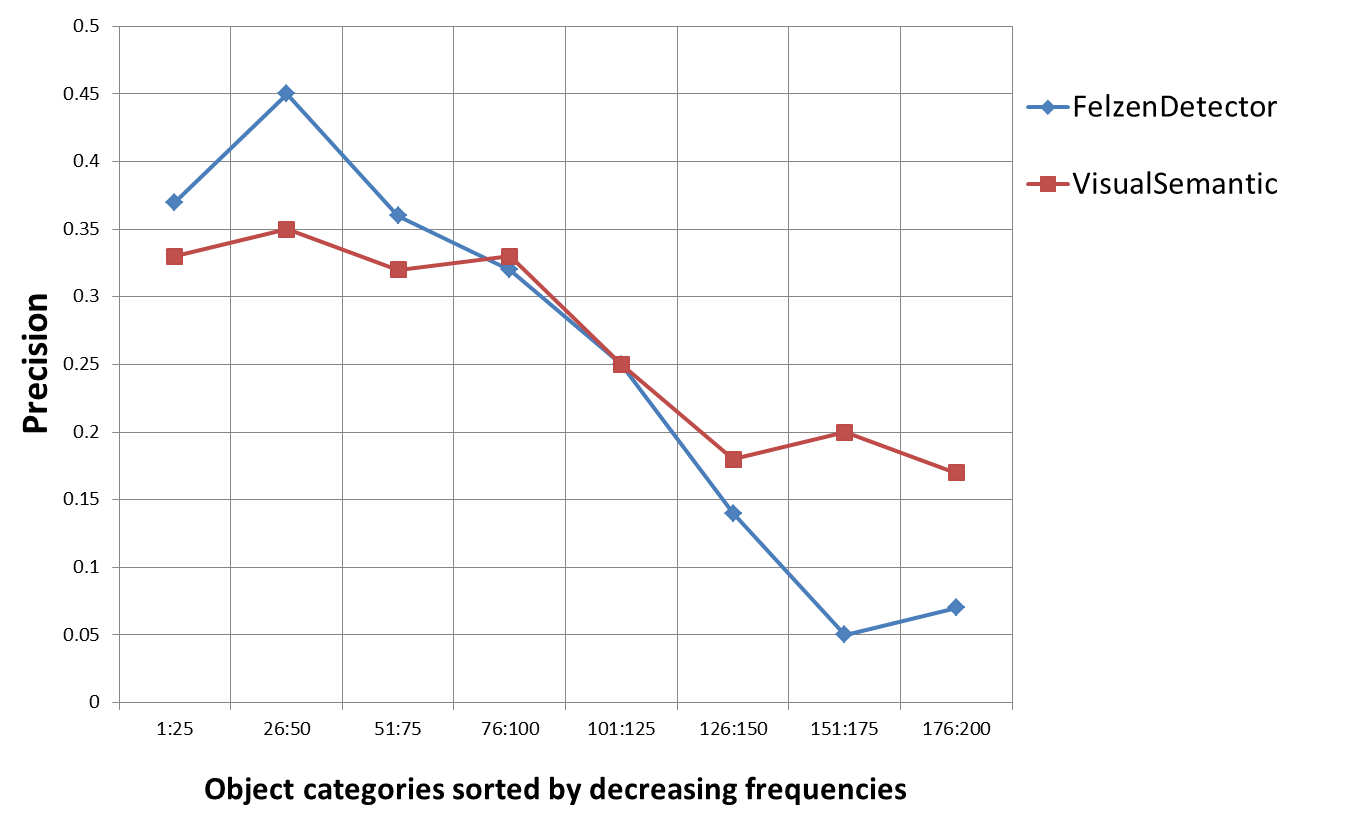}
\caption{Mean precision of object categories sorted by training size. The precisions of every consecutive 25 objects are averaged. Compared to Felzenswalb's DPM detector~\cite{dpm}, our method is much less sensitive to training size.}
\label{fig:objdet}
\end{minipage}
\end{figure*}

We look at the 5 most confident labels predicted by the models and verify their presence in the groundtruth. The results are shown the Figure~\ref{fig:toplabels}. Our performance improves over the hcontext~\cite{hcontext}, both at the initial stage (without semantic context) and after joint inference. This is because, unlike hcontext which relies on DPM outputs and filters out incompatible detections through a tree context, we are able to retrieve missed detections from the visual processing by reinforcing them later through semantics. Table 2 shows results of other baseline generative models. Qualitative results are shown in the supplemental material. 

\begin{table}[t]
\footnotesize
\centering
\begin{tabular}{|l|c|c|c|c|}
\hline
\hline
& TSU &CorrLDA &Init VSIM &Joint VSIM \\
\hline
Scene dist. & 44.03 &33.42 &19.57 &\textbf{13.21} \\
\hline
Top label pred. & 0.29 &0.36 &0.63 &\textbf{0.87} \\
\hline

\end{tabular}
\caption{Model comparisons showing (1) Kullback-Leibler Divergence between estimated and groundtruth scene parameters. (2) Average accuracy of prediction of most confident label in an image.}
\normalsize
\end{table}



\subsection{Object detection:} 
We use precision to report our scores and compare with DPM detector~\cite{dpm} in Figure~\ref{fig:objdet}. The relation between precision and training/learning size is highlighted by sorting object categories from most to least frequent and their precisions are averaged over every 25 objects (for a smooth trend). We see that the DPM precision falls quickly as the size of training set reduces. In contrast, our method generalizes better and performs favorably across all object categories. This advantage on impoverished data is due to a richer set of constraints that prevents overfitting in our model. We show that VSIM better handles the data-imbalance problem frequently seen in many learning problems with natural categories which follow a power law distribution.

We also show the precisions of some objects and compare them to the hcontext detector in Table~\ref{tab:objdet}. We report the precisions at 0.25 False Positives Per Image(FPPI). Since our model can handle fewer training examples, we consider a larger number of object categories (200 vs. 107 in hcontext). The blanks in the table correspond to objects where hcontext gives no response.

\begin{table}[t]
\footnotesize
\centering
\begin{tabular}{|l|l|} \hline
\multicolumn{2}{|c|}{\textit{Precision at .25 FPPI(JointVSIM, InitVSIM, Hcontext~\cite{hcontext})}} \\ 
\hline
\hline
\multirow{2}{*}{a)} & people $(\mathbf{0.74}, 0.0, -)$, cars $(\mathbf{0.68}, 0.42, -)$ \\
& food $(\mathbf{0.63}, 0.0, -)$, picture $(\mathbf{0.71}, 0.25, \mathit{0.76})$ \\
\hline
\multirow{3}{*}{b)} & boat $(\mathbf{0.79}, 0.17, -)$, truck $(\mathbf{.82}, 0.3, \mathit{0.85})$ \\
& painting $(\mathbf{0.62}, 0.23, -)$, poster $(\mathbf{0.55}, 0.0, \mathit{0.57})$ \\
& shop window $(\mathbf{0.73}, 0.11, -)$, balcony $(\mathbf{0.91}, 0.64, \mathit{0.80})$ \\
\hline
\multirow{2}{*}{c)} & videos $(\mathbf{0.93},0.24,\mathit{1.0})$, bottles $(\mathbf{0.72},0.34,\mathit{0.60})$ \\
& books $(\mathbf{0.72},0.28,\mathit{0.80})$, merchandise $(\mathbf{0.93},0.68,-)$ \\
\hline
\multirow{2}{*}{d)} & cow $(\mathbf{0.93},0.0,-)$, fish $(\mathbf{0.87},0.45,-)$ \\
& deck chair $(\mathbf{0.67},0,-)$, umbrella $(\mathbf{0.59},0.07,\mathit{0.57})$ \\
\hline
\hline
\end{tabular}
\caption{Precision scores of some SUN09 objects}
\label{tab:objdet}
\end{table}
\normalsize

\section{Conclusions}
In this paper we have presented VSIM, a scene understanding system that captures both the semantics of a scene and the visual ambiguities that arise due to mapping into image space, within a single model. We explain how VSIM is biologically sound and show that it statistically performs well on a variety of visual tasks. We believe that VSIM maps the lexical-visual space accurately by sharing label hypotheses between semantic and appearance contexts and hence is able to generalize well on new images. In future, we want to develop this method to identify new objects and learn new contexts in the wild.        


{
\bibliographystyle{ieee}
\bibliography{egbib}
}

\end{document}